\documentclass[letterpaper, 10 pt, conference]{ieeeconf}  % Comment this line out if you need a4paper

\IEEEoverridecommandlockouts                              % This command is only needed if 
\usepackage{graphics} % for pdf, bitmapped graphics files
\usepackage{epsfig} % for postscript graphics files
\usepackage{mathptmx} % assumes new font selection scheme installed
\usepackage{times} % assumes new font selection scheme installed
\usepackage{amsmath} % assumes amsmath package installed
\usepackage{amssymb}  % assumes amsmath package installed
\usepackage{algorithm}
\usepackage{algpseudocode}
\usepackage{amsmath,amsfonts}
\usepackage{algorithm}
\usepackage{algpseudocode}
\usepackage{array}
\usepackage[]{subfig}
\usepackage{textcomp}
\usepackage{makecell}
\usepackage{stfloats}
\usepackage{url}
\usepackage{verbatim}
\usepackage{graphicx}
\usepackage{balance}
\usepackage{color, colortbl}
\usepackage{balance}
\usepackage{xcolor}
\usepackage[normalem]{ulem}
\usepackage{booktabs}
\usepackage{tabularx}    % 自适应宽度表格
\usepackage{siunitx}     % 数字对齐
\usepackage{booktabs}    % 专业表格横线
\usepackage{multirow}    % 多行合并

\title{\LARGE \bf
DRARL: Disengagement-Reason-Augmented Reinforcement Learning for Efficient Improvement of Autonomous Driving
Policy
}
\author{Weitao Zhou\textsuperscript{*}, Bo Zhang\textsuperscript{*}, Zhong Cao, Xiang Li, Qian Cheng, Chunyang Liu, Yaqin Zhang, Diange Yang
\thanks{Weitao Zhou, Bo Zhang, Qian Cheng and Diange Yang are with the School of Vehicle and Mobility, Tsinghua University; Bo Zhang and Chunyang Liu are also with Didi Global; Zhong Cao is with Department of Civil and Environmental Engineering, University of Michigan; Xiang Li is with Lab for High Technology, Tsinghua University; Yaqin Zhang is with Institute for AI Industry Research, Tsinghua University, Corresponding to Zhong Cao and Diange Yang (zhcao@umich.edu; ydg@tsinghua.edu.cn)}
\thanks{\textsuperscript{*}These authors contributed equally to this work.}
\thanks{This work is supported by the National Natural Science Foundation of China (NSFC) (52402501, 52394264).}
}

\begin{document}

\definecolor{Gre}{rgb}{0,0.5,0}
\newcommand{\xiang}[1]{\textcolor{blue}{#1}}

\maketitle
\thispagestyle{empty}
\pagestyle{empty}

%%%%%%%%%%%%%%%%%%%%%%%%%%%%%%%%%%%%%%%%%%%%%%%%%%%%%%%%%%%%%%%%%%%%%%%%%%%%%%%%
\begin{abstract}
With the increasing presence of automated vehicles on open roads under driver supervision, disengagement cases are becoming more prevalent. While some data-driven planning systems attempt to directly utilize these disengagement cases for policy improvement, the inherent scarcity of disengagement data (often occurring as a single instances) restricts training effectiveness. Furthermore, some disengagement data should be excluded since the disengagement may not always come from the failure of driving policies, e.g. the driver may casually intervene for a while. To this end, this work proposes disengagement-reason-augmented reinforcement learning (DRARL), which enhances driving policy improvement process according to the reason of disengagement cases. Specifically, the reason of disengagement is identified by a out-of-distribution (OOD) state estimation model. When the reason doesn’t exist, the case will be identified as a casual disengagement case, which doesn’t require additional policy adjustment. Otherwise, the policy can be updated under a reason-augmented imagination environment, improving the policy performance of disengagement cases with similar reasons. The method is evaluated using real-world disengagement cases collected by autonomous driving robotaxi. Experimental results demonstrate that the method accurately identifies policy-related disengagement reasons, allowing the agent to handle both original and semantically similar cases through reason-augmented training. Furthermore, the approach prevents the agent from becoming overly conservative after policy adjustments. Overall, this work provides an efficient way to improve driving policy performance with disengagement cases.

\end{abstract}

\section{Introduction}

In recent years, an increasing number of autonomous vehicles (AVs) have been deployed on open roads, achieving remarkable performance milestones, such as an average of 47,000 km per disengagement in US \cite{sinha2021crash}. However, AVs can still encounter failures during real-world driving, leading to driver-initiated disengagements.

The majority of disengagement cases stem from driving policy failures, highlighting areas where the policy needs improvement. In theory, data-driven driving policies—such as those powered by reinforcement learning agents \cite{lin2020anti, zhou2023identify}—should be able to enhance themselves automatically using disengagement data. However, directly training these policies with raw disengagement data does not consistently lead to improved performance. This is due to several challenges: (1) The root cause of a policy failure may not occur at the moment of disengagement but at an earlier stage; (2) The available disengagement data is often insufficient for effective policy training \cite{wang2020generalizing}; and (3) Not all disengagements are the result of policy failures—some, such as casual driver interventions, should be excluded from the training process as they do not reflect deficiencies in the driving policy.

\begin{figure}[!t]
\centering
\includegraphics[width=\linewidth]{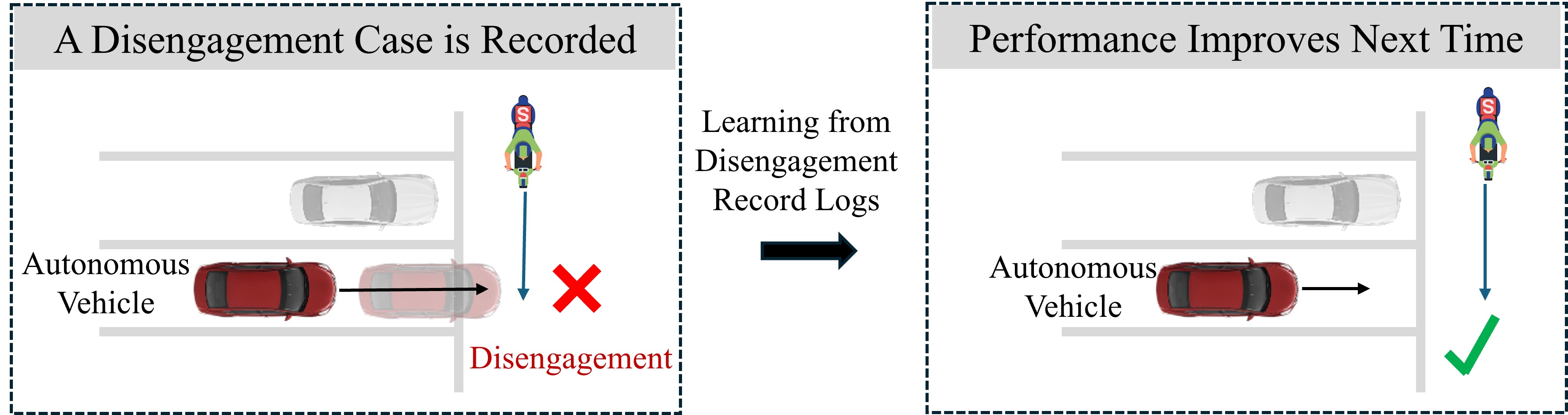}
\caption{The autonomous vehicle is intervened by driver because of potential driving policy failure in the left figure. The goal of this work is to find the reason causing disengagement and improve the driving policy when encountering similar scenarios in the future.}
\label{fig:introduction}
\vspace{-5mm}
\end{figure}

\textit{To this end, this work proposes the idea of automatically improving the driving policy according to the reasons for disengagement in addition to original data.} 
Our main idea is to first determine the reason of disengagement and then adjust the driving policy using reason-augmented data. In this work, we specifically consider disengagement cases caused by abnormal behavior of surrounding objects, which results in unreasonable driving performance.
To trace the reason for disengagement, we utilize historical training data to identify out-of-distribution (OOD) states that the driving agent has not encountered before. If a disengagement case lacks a traceable reason, it is classified as an case not caused by the policy’s failure.
Once the reason is identified, the original disengagement data is used to construct a corresponding reason-augmented imagination environment. This environment, as illustrated in Fig. \ref{fig:introduction}, leverages driving models to generate diverse training data while preserving the identified disengagement reason.

The purpose of this work is that when an autonomous vehicle performs unsatisfactorily and is intervened during road testing, the proposed method can support the driving agent to automatically learn from this case and perform better when meeting similar cases in the future.
As a continuation of our former publishing \cite{cao2022autonomous}, this work proposes a better method to explicitly locate the reason of disengagement and refines the building of reason-augmented imagination environment.
The detailed contribution is as follows:

\begin{enumerate}

    \item Automated Disengagement Reason Identification: We propose a method for identifying policy-related disengagement reasons through out-of-distribution (OOD) state detection, eliminating the need for expert annotations or customized datasets, providing a scalable solution for disengagement reason analysis. 
    \item Reason-Augmented Imagination Environment: We proposes a reason-augmented imagination environment, which combines real-world driving logs and parametric scenario variations considering the moment disengagement reason happens. This environment enables the policy to improve itself using disengagement cases.
    \item     Disengagement-Reason-Augmented Reinforcement Learning (DRARL) framework: The framework enables autonomous driving agents to automatically learn from disengagement cases. By incorporating disengagement reason analysis into the learning process, the framework improves the agent's performance when encountering similar scenarios in the future. 

\end{enumerate}

% The structure of the whole manuscript is as follows. 
% Section II introduces the related works. 
% Section III formally defines the problem and introduces the automatic learning framework. 
% The disengagement reasoning method is described in section IV. 
% Section V introduces the disengagement case augmentation method.
% Section VI describes the experiments and Section VII concludes the whole work. 

\section{Related Works}

\subsection{Driving Policy Failure Reasoning}

This work involves finding the reasons why RL agent fails in certain driving scenarios.
Existing works implement the reasoning of failure either from the input of the agent, which is the environmental state or from the output of the agent.

In the category of reasoning from the state, some representative works try to identify risky objects in driving environments and take them as the potential reason for failure.
Researchers from \cite{li2020make} present a two-stage risk object identification framework based on causal inference with the proposed object-level manipulable driving model.
In \cite{gao2019goal}, researchers proposed a framework that incorporates both visual model and goal representation to conduct object importance estimation.
In \cite{zeng2017agent}, researchers realize the risky region localization by explicitly
modeling both spatial and appearance-wise non-linear interactions between the agent triggering the event and another agent involved.
These referred methods require large-scale datasets, while disengagement cases are usually too rare to form one.

In another category of reasoning from the output of the agent, researchers usually append accessory modules to the RL agent to assess the probability of inference failure. 
In \cite{kahn2017uncertainty}, researchers proposed a model-based learning algorithm that estimates the probability of collision with a statistical estimate of uncertainty. In \cite{cao2021confidence}, the true action value function distributions are calculated based on Lindeberg-Levy Theorem, and the RL agent is considered as low confidence when its action value is lower than the baseline policy. In a model-based RL method, the uncertainty of the agent is associated with inaccurate model error, policy deviation error, and model rollout steps  \cite{wu2022uncertainty}\cite{YangWGYJLW22}. 
High uncertainty indicates the high failure possibility of the RL agent \cite{YangWWL24}.
The aforementioned methods are closely related to the RL agent's implementation, while we don't expect the proposed method to have any assumptions on driving policy in this work.

Based on the above considerations, this work reasons from the state and defines the reason for disengagement as a period of behaviors that the driving policy seldom met before.

\subsection{Imagination Environment from Real Cases}

This work intends to build a imagination environment according to the disengagement cases, which belongs to the real data enhanced simulation domain. 
A basic but practical idea is to directly replay the recorded trajectories of all the surrounding objects in a simulator, where the RL agent can learn a better policy.
Such an idea is widely used in accident analysis, e.g., in the German In-depth Accident Study (GIDAS) \cite{fagerlind2010analysis}, and National Highway Traffic Safety Administration (NHTSA) \cite{cerf2018comprehensive}. 
However, the surrounding objects in this environment cannot react to the ego vehicle's policy change. 
It makes the generated policy in this environment overfit to a specific trajectory, which may be more dangerous.

Another idea is to form an interactive environment by adding behavior models to the environment objects. 
In these methods, the real data serves to provide the initial state of the simulation.
There has been various research to build the behaviors for vehicles or pedestrians. For example, the intelligent driver model (IDM) \cite{treiber2000congested} and the minimizing overall braking induced by lane change (MOBIL) model \cite{kesting2007general} have been widely used to simulate the traffic environment.
Some recent research tries to use machine learning models to improve the interaction ability of the models \cite{suo2021trafficsim}\cite{bergamini2021simnet}.
However, if the surrounding objects change their trajectories, 
likely, they won't perform the behavior which causes the disengagement.

This work will use the reason for disengagement to enhance the constructed imagination environment. It combines the reason and driving models to provide reactive but relevant data.

\section{Problem Definition}

\begin{figure}[!b]
\centering
\includegraphics[width=\linewidth]{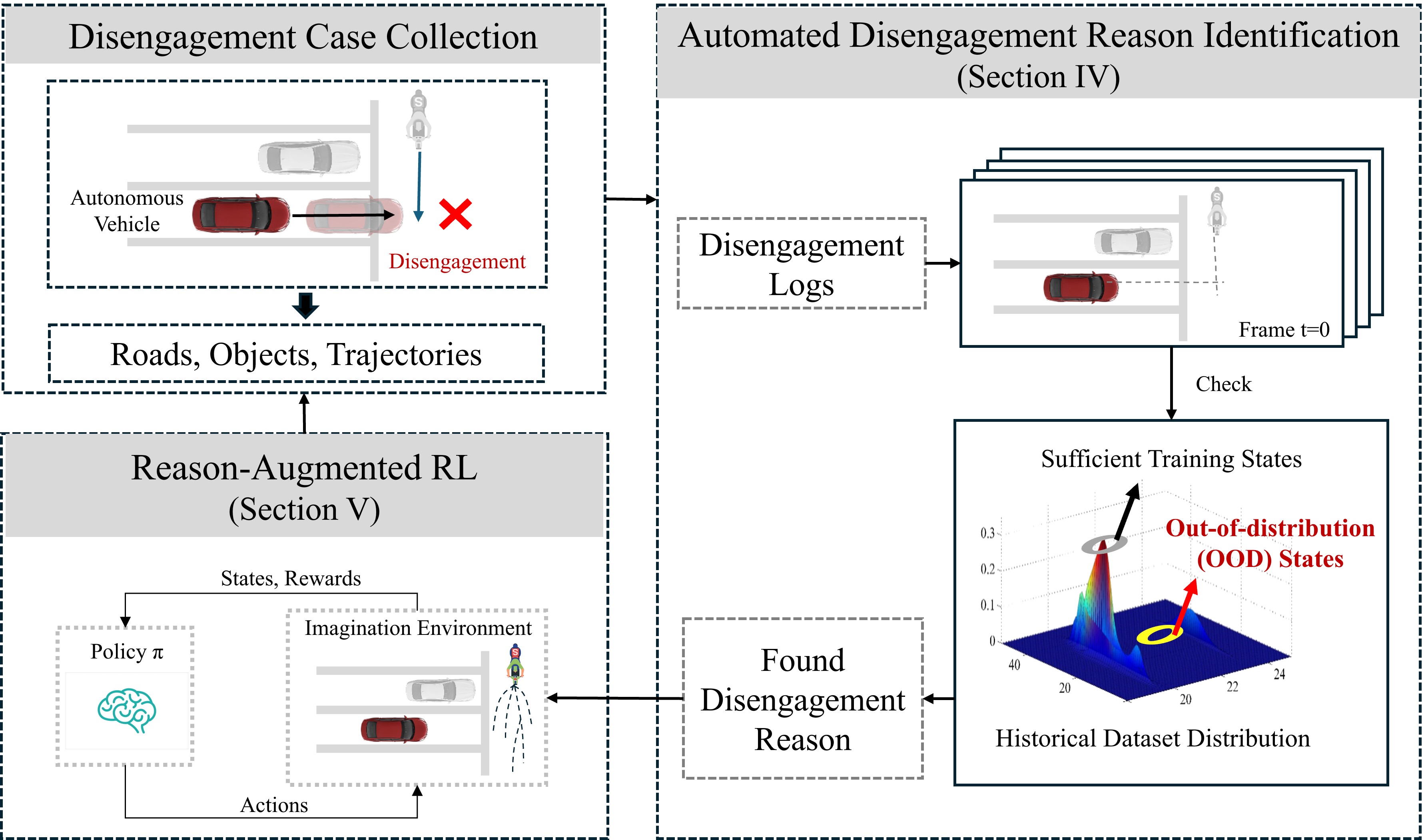}
\caption{The framework of disengagement reasoning for policy improvement.
}
\label{fig:framework}
\end{figure}

The proposed method works for the following problem:
\textit{An autonomous vehicle drives on the road and is intervened by the drivers.
The AV then record the trajectory and the environment within seconds before the disengagement, as a disengagement case.
Our method should automatically learn to handle this case and to get better performance in the similar cases in future driving.}
The concrete details for achieving this are as follow:

For each disengagement case $\theta$ in collected disengagement case set $\Theta$, it's defined as a sequence of observational states $\{s^0, s^{1}, ..., s^a\}$ from timestamp $0$ to $a$. 
The state of each frame is expressed as $s = \{q_e, q_1, q_2, ..., q_n\}$, where $q_e$ is the kinematic state of ego vehicle and $q_1, q_2, ..., q_n$ are the set of kinematic states of $n$ observed surroundings.
The HD map $M$ where the disengagement cases are collected is also available.
This work will firstly analyze the reason for disengagement, which is expressed as a set of surroundings' state sequences $R$. The element in this set $q_i^{b:d}$ denotes the state sequence of object $i$ from timestamp b to d.
If this set is empty, this case is considered casual disengagement and deprecated.
If not, the reason $R$ and the disengagement recording $\theta$ will be used to build an imagination environment $E(R, \theta)$. The driving agent will run in this environment and improve its ability on $\theta$.
The framework is shown in Fig. \ref{fig:framework}.

\section{Disengagement Reason Identification }

\subsection{The Definition of Disengagement Reason}

\subsubsection{Definition}

Considering the fact that most disengagement cases caused by driving policy failure are due to dynamic objects in the scenarios \cite{sinha2021crash}, this work defines the reason of disengagement $R$ as a set of surroundings’ state sequences from the recording:

\begin{equation}
R = \{q_i^{b:d}\}.
\end{equation}

From this definition, we can express the reason more concretely as the the start and end timestamp $b$ and $d$ of single or several objects, which is indexed by $i$. 

\subsubsection{Start of Disengagement Reason}

Defining the start of a disengagement reason requires identifying when the driving policy is likely to fail. Reinforcement learning (RL) has shown strong performance in well-defined scenarios, even surpassing humans in tasks where they traditionally excel \cite{holcomb2018overview}. However, RL struggles to generalize beyond its training distribution \cite{zhou2023identify,abs-2302-09601}. While RL agents perform well in environments with fixed state distributions, they often fail when encountering out-of-distribution (OOD) states, even in diverse training environments \cite{cobbe2019quantifying}.

This limitation is critical for autonomous driving, where vehicles operate in highly dynamic environments. A single inference failure by the RL agent can lead to a chain of states ending in unavoidable failure. Based on these observations, this work defines the occurrence of an OOD state as the start of a disengagement reason. 

\subsubsection{End of Disengagement Reason}

In this work, we take the timestamp of driver disengagement as end timestamp $d$.
The consideration here is that the reason causing driver to intervene the vehicle is the performance of AV didn't meet his expectation.
Consequently, when the driver intervened the AV, the reason must have happened and already caused the performance problem of AV agent.
After perceiving the performance problem, the driver will choose to intervene the vehicle.
Therefore, the disengagement time is proper for marking the end of reason.

\subsection{Out-of-distribution States Identification}

% The target of out-of-distribution states detection is to realize a function $F(t, i, \theta)$, which leverages the recording $\theta$ to check if the state of certain object $q^t_i$ is out of distribution. 

It's hard to directly model the distribution of states because the state space is continuous and high-dimensional.
Consequently, this work will leverage the advance in generative model and realize detection by checking if the states in disengagement cases can be generated by a model summarizing all historical data.
This approach involves two parts, the state  generation model and similarity measurement.

\subsubsection{Historical Trajectory Generation}
This work formulates the state generation problem as a motion prediction problem, whose target is to predict the future state given current and past states, expressed as:

\begin{equation}
O_i^{t+\gamma} = \mathbf{N}_D(s^{t-c:t}),
\end{equation}
where $O_i^{t+\gamma}$ is the set of possible future location of object $i$ after $\gamma$ frames. 
$c$ is the former frames that the model needs to consider.
$\mathbf{N}_D$ is the prediction model trained on RL agent's historical trajectory set $D$.
This nature makes our approach available only on RL agent with accessible replay buffer.

Our approach requires a nondeterministic predictor to generate multi-styled and diversified trajectories once happened in RL agent's history.
Here our model refers to the model design in \cite{salzmann2020trajectron++}, leveraging Conditional Variational Auto-Encoder (CVAE) and Gaussian distribution over output to realize styled and nondeterministic prediction.

The CVAE is to explicitly realize the multimodality of predicted states by introducing a discrete categorical latent variable $z$.
The discrete latent variable $z$ encodes high-level latent behavior style and transforms the prediction target $p(o_i^{t+\gamma}|s^{t-c:t})$ to the conditioned version:

\begin{equation}
p(o_i^{t+\gamma}|s^{t-c:t}) = \sum_{z \in Z}p_\phi(o_i^{t+\gamma}|s^{t-c:t}, z)p_\epsilon(z|s^{t-c:t}),
\end{equation}
where $\phi,\epsilon$ are respectively deep neural network parameters.

The outputs of prediction networks are parameters of bivariate Gaussian distribution over steering angle and acceleration.
% , which provides more real results than deterministic version.
The action distribution will be sampled and then used for generating location after $\gamma$ frames through a dynamic integrator.

\subsubsection{Similarity Measurement}

Given the trajectory generation model, we can sample batches of future states to represent the states in historical data.
The target here is to measure the similarity between the real state and the generated states.
The proposed method is to fit the generated states into a distribution, and calculate the probability that the real state belongs to this distribution, expressed as:

\begin{equation}
e = p(\Tilde{q}^{t+\gamma}_i \in K(O_i^{t+\gamma})),
\end{equation}
where $e$ denotes the probability.
$\Tilde{q}^{t+\gamma}_i$ denotes the location of object $i$ after $\gamma$ frames. 
$K$ denotes the distribution estimation from generated locations.
This work adopts the Kernel Density Estimation \cite{parzen1962estimation} as the distribution estimation method, which is a non-parametric estimation method and can fit complex distributions.
If the probability $e$ is lower than a hyperparameter $e_o$, the state at timestamp $t$ is considered out of distribution.

\subsection{Disengagement Reasoning}

\begin{figure}[!t]
\centering
\includegraphics[width=\linewidth]{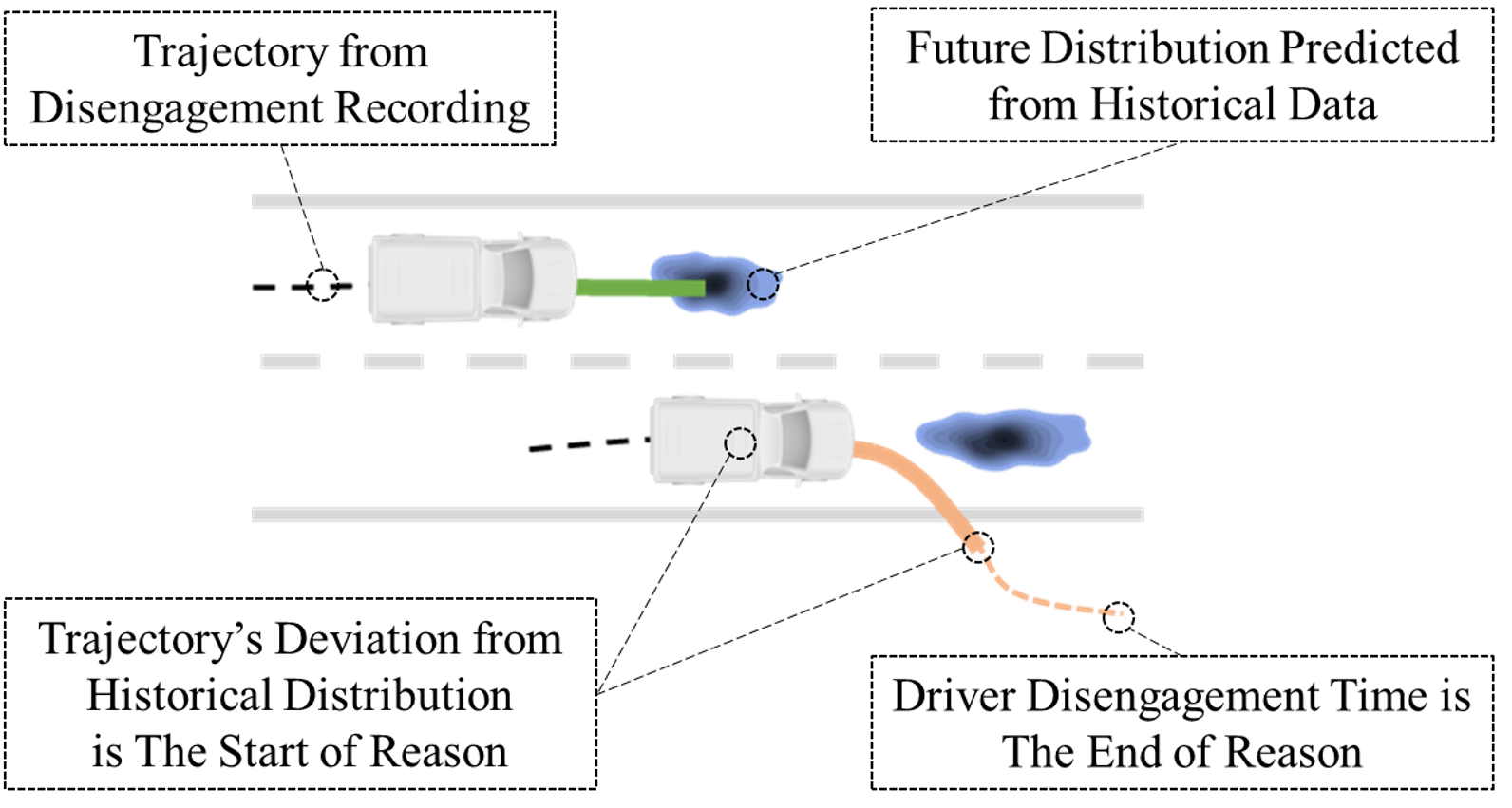}
\caption{The disengagement case reasoning method.
}
\label{fig:method}
\end{figure}

This section introduces the specific steps for finding the disengagement reason $R$.
Based on the definition, to find the reason means to find the start and end timestamp $b$ and $d$ for each object in recording if it performed out-of-distribution states.

The overall method is demonstrated in Fig.\ref{fig:method}.
The disengagement time is extracted from recording as the end timestamp $d$ of the potential reason.
From end timestamp $d$, we trace back to find the start timestamp $b$ until the start of recording.
The process is iterated over each object in the disengagement recording.
The start timestamp of certain object is the time of its first out-of-distribution state, which is checked by the method introduced in the last section. 
If the method doesn't find any out-of-distribution state, this case is classified as the casual disengagement.

\section{Reason-Augmented Reinforcement Learning}

\subsection{Reason-Augmented Imagination Environment Formulation}

 Imagination means that the environment does not exist but can be considered as the data provider for policy training. 
Thus, designing the imagination environment should form all the elements in an environment.

The imagination state is defined as follows:
\begin{equation}
\label{eqn:imagination state}
s = \{q_e, V_e, \{q_i, V_i\}\},
\end{equation}
where $q_e=\{x_e, \dot{x}_e, y_e, \dot{y}_e\}$ and $q_i=\{x_i, \dot{x}_i, y_i, \dot{y}\}$ denotes respectively the kinematics of ego vehicle and all the objects in the scenario.
% Compared with the state formulation in disengagement recording, the state in imagination environment add a transition model to each object in the environment.
$V_e$ denotes the transition model of the ego vehicle, considering the kinematics $q_e^t$ and the taken action $a^t$ of the ego vehicle. 
Namely, $q_e^{t+1}=V_e(q_e^t, a^t)$. 
The ego vehicle transition model is provided by the simulator. 
The action taken by RL agent $a^t$ only depends on the current state $s^t$, namely, $a^t = RL(s^t)$.
$V_i$ denotes the transition model of each surrounding object, considering its current state and the ego vehicle's state.
Namely, $q_i^{t+1}=V_i(q_i^t,q_e^t)$.

This work adopts the Markov Decision Process (MDP) \cite{kochenderfer2015decision} to model this environment.
Following the assumption in MDP, the transition model of environment can be defined as:
\begin{equation}
\label{eqn:imagination_transition}
s^{t+1} = \mathcal{P}(s^t,a^t) = \{V_e(q_e^t, RL(s^t)), V_e, \{V_i(q_i^t,q_e^t), V_i\}\},
\end{equation}
where $\mathcal{P}$ denotes the transition model in the MDP problem.

From Eq. \ref{eqn:imagination_transition}, we can find the next state $s^{t+1}$ in imagination environment is only relevant with current state $s^t$ and surroundings' transition model $V_i$.
Consequently, an imagination environment can be generated by defining the initial state $s^0$ and surroundings' transition model $V_i$.

This work chooses the state at the time that agents in the reasoning set R begin to appear as the initial state of the imagination environment.
This choice guarantees that the object causing disengagement will always be present in the imagination environment.
The surroundings' transition model is further designed in the next section.

\subsection{Reason-Augmented Transition Model}

The target of imagination environment is to improve the driving performance for the specific reason causing disengagement, while providing diversified data to avoid overfitting to a specific scenario.
To this end, this work uses the reason to enhance the transition model as follows:

\begin{equation}
\label{eqn:transition_model}
V_i = \{(V^r_i|t < b), (V^m_i|t \geq b)\} 
\end{equation}
%or \{(V^r_i|t < d), (V^m_i|t \geq d)\}
where $V_i$ denotes the driving model of the $i^{th}$ object and $t$ denotes the timestamp in the imagination environment. 
$V^r_i$ denotes the log-replay model, which only follows the recorded trajectory of the $i^{th}$ object. 
$V^m_i$ denotes an interactive model, which can react to the ego vehicle's driving states and produce various behavior.
$b$  denotes the beginning of reason for object $i$.

In this way, $V_i$ has switches between two modes. One focuses on increasing the data diversity after the reason starts. In this mode, the reason is very likely to not reappear. It helps to avoid overfitting to states before reason.
Another mode focus on the data after the reason ends. In this mode, the reason is reappeared. Driving policy can learn to handle the reason while avoiding overfitting to states after the reason.
These two modes are randomly chosen in each training episode.

The log-replay model $V^r$ is defined as follows:
\begin{equation}
\label{eqn:replay_model}
V^r_i: q^{t}_i = \theta_i^t
\end{equation}
where $\theta_i^t$ denotes the recorded trajectory of $i^{th}$ object at time $t$.
The log-replay model can perfectly reproduce the same scenario as disengagement. However, it's fixed and cannot react to the ego policy changing.
The objects not in reason set uses only the log-replay model.

The interactive driving model $V^m_i$ uses some principle-based model with the random noise, defined as follows:
\begin{equation}
\label{eqn:principle_model}
V^m_i: q^{t+1}_i = f(q^t_e, \{q^t_i\}),
\end{equation}
where $f$ is a behavior model according to the attribute of the object, which generates the next state according to all objects' states in the environment.
% If the $i^{th}$ is a vehicle, then $f$ could be the driver models.
% If the $i^{th}$ is a walker, then $f$ could be the pedestrian walking model. 
There are various works summarizing the existing interactive models \cite{bevly2016lane, xiao2010comprehensive, teknomo2006application}.
More realistic models can make the improved policy approach the optimal policy in real-world driving.
This work does not limit to a specific interactive model and better interactive models may further improve the performance.

\section{Evaluation}

% This evaluation tries to show the advantage of proposed method in two aspects. Firstly, the proposed method can distinguish casual disengagements from disengagement recordings.
% For the disengagement caused by policy failure, this method can locate the specific object and its behavior causing disengagement.
% Then, knowing the reason, the driving policy can get improved with the reason-augmented imagination environment.

\subsection{Evaluation Case Setting}

This section will introduce the disengagement cases design for evaluation, the original policy design before improvement, baseline method and performance metrics.

\subsubsection{Real-world Disengagement Cases for Evaluation}

This study aims to improve autonomous driving policies in disengagement cases. We leverages large-scale real-world operational data collected from autonomous vehicles during on-road testing across urban areas in Beijing, Guangzhou from China. 

\begin{figure}[!b]
\centering
\includegraphics[width=\linewidth]{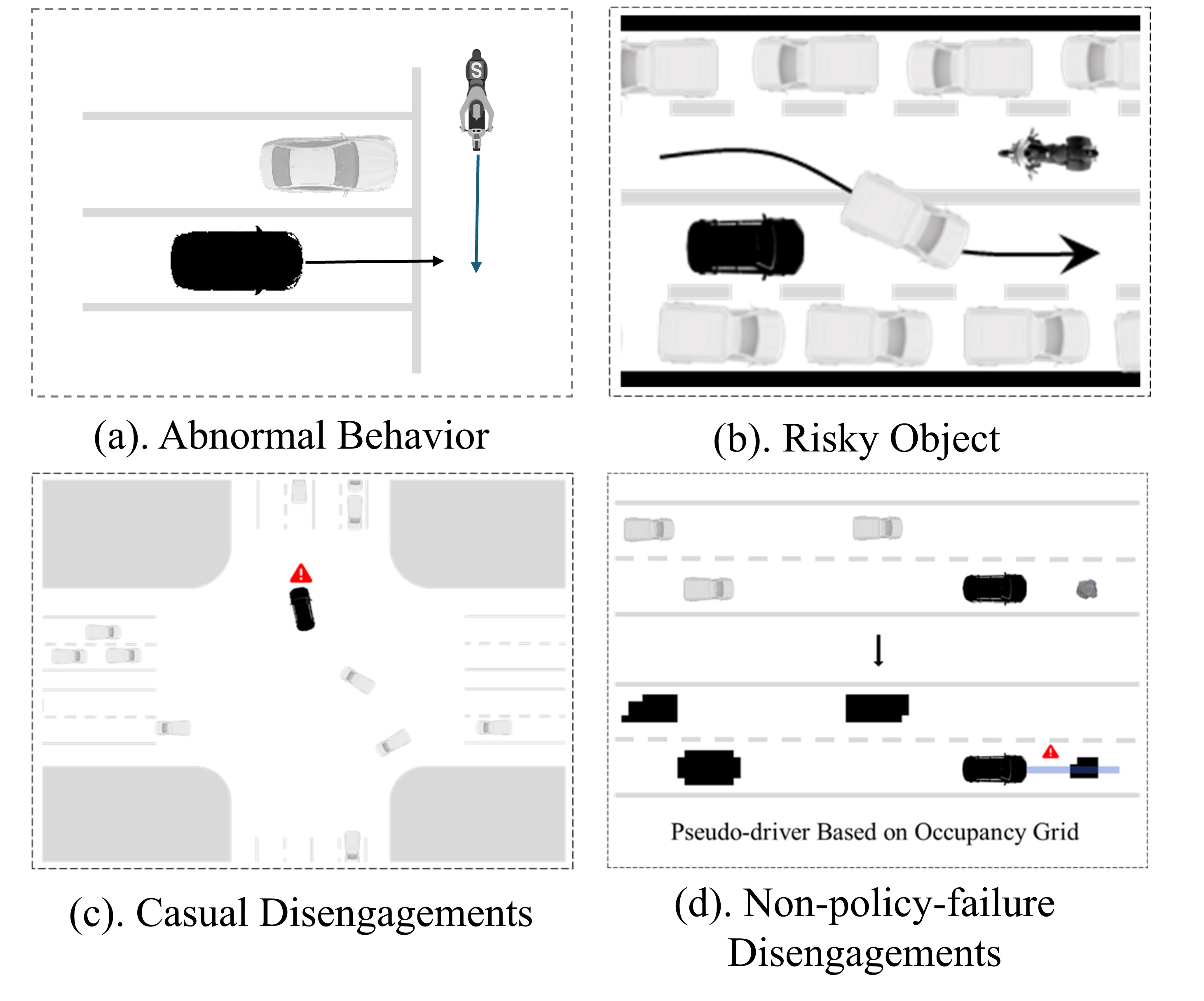}
\caption{Collected Disengagement Cases for Evaluation.}
\label{fig:scenarios}
\end{figure}

Through systematic analysis of 2,843 disengagement events, we categorize critical disengagement scenarios into four distinct classes and reconstruct representative cases in CARLA simulator with desensitization methods, as shown in Fig. \ref{fig:scenarios}.

\textbf{Abnormal Behavior Case:} The first type of case is a cyclist or pedestrian rushing across the road, as shown in Fig. \ref{fig:scenarios}(a).  This cyclist(pedestrian) wants to speed through the intersection regardless of oncoming ego vehicle, producing a dangerous situation. We reproduces 100 cases in this type from real-world disengagement cases, with various environmental object trajectories, i.e., the cyclist(pedestrian)'s position and speed profile.
Through this scenario, this work tries to validate if the proposed method can locate this abnormal behavior and improve policy correspondingly.

\textbf{Risky Object Case:} Another scenario is a sudden overtaking as shown in Fig. \ref{fig:scenarios}(b).
In this case, ego vehicle is driving on a two-way single-lane road.
There are various parking vehicles on the roadsides. 
A social vehicle tries to overtake our AV crossing the double yellow line. 
Since there may be objects moving on the opposite lane, this vehicle takes a very emergency cut-in to drive back to the lane.
We reproduces 100 cases in this type from real-world disengagement cases. Through this scenario, this work tries to evaluate if the proposed method can find the vehicle causing disengagement among vehicles, and locate its concrete behavior.

\textbf{Casual Disengagement Cases:} This experiment tries to evaluate whether the proposed method can identify casual disengagement cases as cases that have no disengagement reasons, preventing unnecessary policy training.
To achieve this, this work samples 2400 frames from normal driving data and assumes that every frame is a casual disengagement case. 
One example frame is shown in Fig. \ref{fig:scenarios}(c).
The rate of a case identified as without reason is evaluated.

\textbf{Non-policy-failure Disengagement Cases:} In certain situations, a driver might take over due to problems unrelated to the driving policy itself, e.g., obstacles on roads that are neglected by the perception system.
This experiment tries to evaluate whether the proposed method can identify non-policy-failure disengagement cases as cases that have no disengagement reasons pertaining to the policy itself, preventing unnecessary policy training.
To achieve this, this work places garbage in AVs' routes in each running.
We deploy a planner based on the occupancy grid \cite{lu2014layered} which can recognize the garbage as obstacles to simulate a human driver.  
Every time the pseudo-driver makes a stop is considered a non-policy-failure disengagement, then the running is restarted.
We repeat the running in the simulated environment and collect 600 frames for evaluation. One example is shown in Fig. \ref{fig:scenarios}(d).
The rate of a case identified as without reason pertaining to the policy is evaluated.

\subsubsection{Test Cases for Improved Policy}

For the first and second type of scenario, this work will use the found reason to build a reason-augmented imagination environment.
The original policy will be trained in this environment to improve its ability.
The interactive driving models $V^m_i$ choose those used in original policy training.

We randomly select less than 5\% of disengagement cases in each type of scenario for training. 
The testing cases, designed to evaluate reasoning and policy improvement capabilities, are derived from real-world disengagement data and categorized into two types based on distinct testing objectives. 
The first objective is to assess whether the improved policy exhibits overly cautious behavior toward surrounding objects when they do not intend to act aggressively. Overly cautious behavior indicates that the agent fails to improve based on the correct disengagement reasons and instead becomes excessively conservative.
This improvement may largely affect the normal driving efficiency, which is unsatisfactory. 
To test this, in Scenario 1, the pedestrian or cyclist performs a slight acceleration as the autonomous vehicle (AV) approaches the intersection. In Scenario 2, the surrounding vehicle executes a minor lateral movement within its lane as it approaches the AV.  

Another group of test cases is designed to evaluate whether the improved policy can successfully handle similar disengagement scenarios without introducing perturbations, ensuring it remains effective within the same category. 

% The specific parameters of 2 groups of cases are shown in the Appendix.

\subsection{RL policy and Baselines Design}

\subsubsection{Original Policy before Disengagement}

We pre-trained a policy using RL as the original policy for improvement in the CARLA simulator \cite{dosovitskiy2017carla}.
The training scenarios have the same map with the testing cases. In the training process, a certain amount of vehicles and pedestrians are randomly spawned in the simulator, heading for random locations.

The formulation of RL policy is as follows:

\textbf{State Space:} The definition of state space follows Eq. \ref{eqn:imagination state}, while the transition models are unobserved to simulate real-world situations. 

\textbf{Action Space:} The action space of RL agent is defined as:
\begin{equation}
    \label{eqn:action_space}
    a \in \mathcal{A}, a = \{\dot{x}_e, \dot{y}_e\},
\end{equation}
where $\dot{x}_e,\dot{y}_e$ denotes the target longitudinal and lateral acceleration.

\textbf{Reward Function:} The reward encourages higher driving efficiency, making progress to pass the scenario, and avoid collisions, which is defined as follows:
\begin{equation}
    r(s)=r_e+r_p+r_c
\end{equation}
where $r_e$ denotes efficiency reward proportional to the ego vehicle velocity, $r_p$ denotes progress reward proportional to the ego vehicle's progress along reference line , $r_c$ denotes collision penalty.

% \begin{equation*}
% {r(s)} = \begin{cases}
% {-100}, &{\text{if}\ s \in \mathcal{D}}  \\ 
% {\dot{x}_e/8 - y_e}, &{\dot{x}_e < 8} \\
% 0, &{\text{otherwise,}}
% \end{cases}
% \end{equation*}
% where $\mathcal{D}$ denotes the terminal state sets, including collision and long-term stillness.

% The surrounding vehicles adopt IDM for longitudinal decisions and MOBIL for lateral decisions, with parameters shown in Table. \ref{tab:model_param}.
% The pedestrians are controlled by a constant velocity path follower.

% The model parameters of IDM and MOBIL are listed in the following table.

% \begin{table}[h]
% \centering
% \caption{Principle-based Model Parameters}
% \begin{tabular}{
% >{\arraybackslash} m{3cm}
% >{\centering\arraybackslash} m{1.6cm}
% >{\centering\arraybackslash} m{2.6cm}
% }
%  \Xhline{4\arrayrulewidth}
%  Parameters& Symbol & Value\\
%  \hline
%  Desired velocity & $\dot{s}_0$ & 8$\times\mathcal{N}(1, 0.5^2)$ m/s \\
%  Exponent for velocity & $\theta$ & 4 \\
%  Desired time gap & $T$ & 1.5s \\
%  Jam distance & $g_0$ & 2.0$\times\mathcal{N}(1, 0.5^2)$m \\
%  Max acceleration & $a$ & 1.4$\times\mathcal{N}(1, 0.5^2)$$m/s^2$\\
%  Desired deceleration & $b$ & 2.0$m/s^2$ \\
%  Politeness & $p$ & 0.5$\times\mathcal{N}(1, 0.5^2)$ \\
%  Lane change threshold & $\Delta a$ & 0.1 \\
%  \Xhline{4\arrayrulewidth}
% \end{tabular}
% \label{tab:model_param}
% \end{table}

\subsubsection{Baseline and Metrics}

For the casual and non-policy-failure disengagement identifying problem, this work takes the successful identifying rate as the metric, expressed as $\epsilon = n_p / n_a$,
where $\epsilon$ is the passing rate, $n_p$ is the number of frames recognized as casual disengagement. $n_a$ is the total frames.
% \begin{equation}
% \label{eqn:reason_rate}
% \epsilon = n_p / n_a,
% \end{equation}

For the subsequent policy improvement in reason-augmented environment, this evaluation takes a widely-used Soft Actor-Critic based \cite{haarnoja2018soft} RL agent as the first baseline. It's trained by directly replaying the log recording in imagination environment without the related reasoning processing introduced in this work. 

For comparison, we also established Randomly Reason Reinforcement Learning as a baseline. This method is identical to our proposed DRARL in all aspects except one: it randomly selects the start of reason moment for disengagement cases during training. Furthermore, we evaluate the performance of a policy trained similarly to DRARL but with manually predefined fixed start moments for reasoning (distinct from the moments identified by DRARL), referred to as Fixed Reason-Augmented Reinforcement Learning (FRARL). 

This work takes the collision rate and passing rate as the driving performance. The passing rate can be calculated as $\lambda = n_g / n_t$,
where $\lambda$ is the case passing rate, $n_g$ is the number of passed scenarios (without collisions or stucks), and $n_t$ is the number of all derive cases. The collision rate can be calculated is similar way.
% The evaluation metric is thus the improved performance from the original policy $\pi_0$ to trained policy $\pi$, namely, $\lambda_\pi - \lambda_{\pi_0}$. 

\begin{table}[!t]
\centering
\caption{The Results of Identifying Rate}
\begin{tabular}{
>{\arraybackslash} m{4.5cm}
>{\centering\arraybackslash} m{3.1cm}
}
 \Xhline{2\arrayrulewidth}
 Scenario & Identifying Rate\\
 \hline
 Casual Disengagements &  92.5\% (2221 of 2400) \\
 Non-Policy-Failure Disengagements & 94.7\% (568 of 600)\\
 \Xhline{2\arrayrulewidth}
\end{tabular}
\label{tab:rl_param}
\end{table}

\subsection{Result for Disengagement Case Reasoning}

We used the proposed OOD states detection method to check the reason for disengagement. 
For each object, we sample 1000 futures states from the generative model to fit the state distribution.
The time between each frame is 0.5 seconds, and the hyperparameter $\gamma$ is set to 2 to simulate the human reaction time.
The possibility threshold $e_o$ is 0.01\%.

\begin{figure}[!t]
\centering
\includegraphics[width=\linewidth]{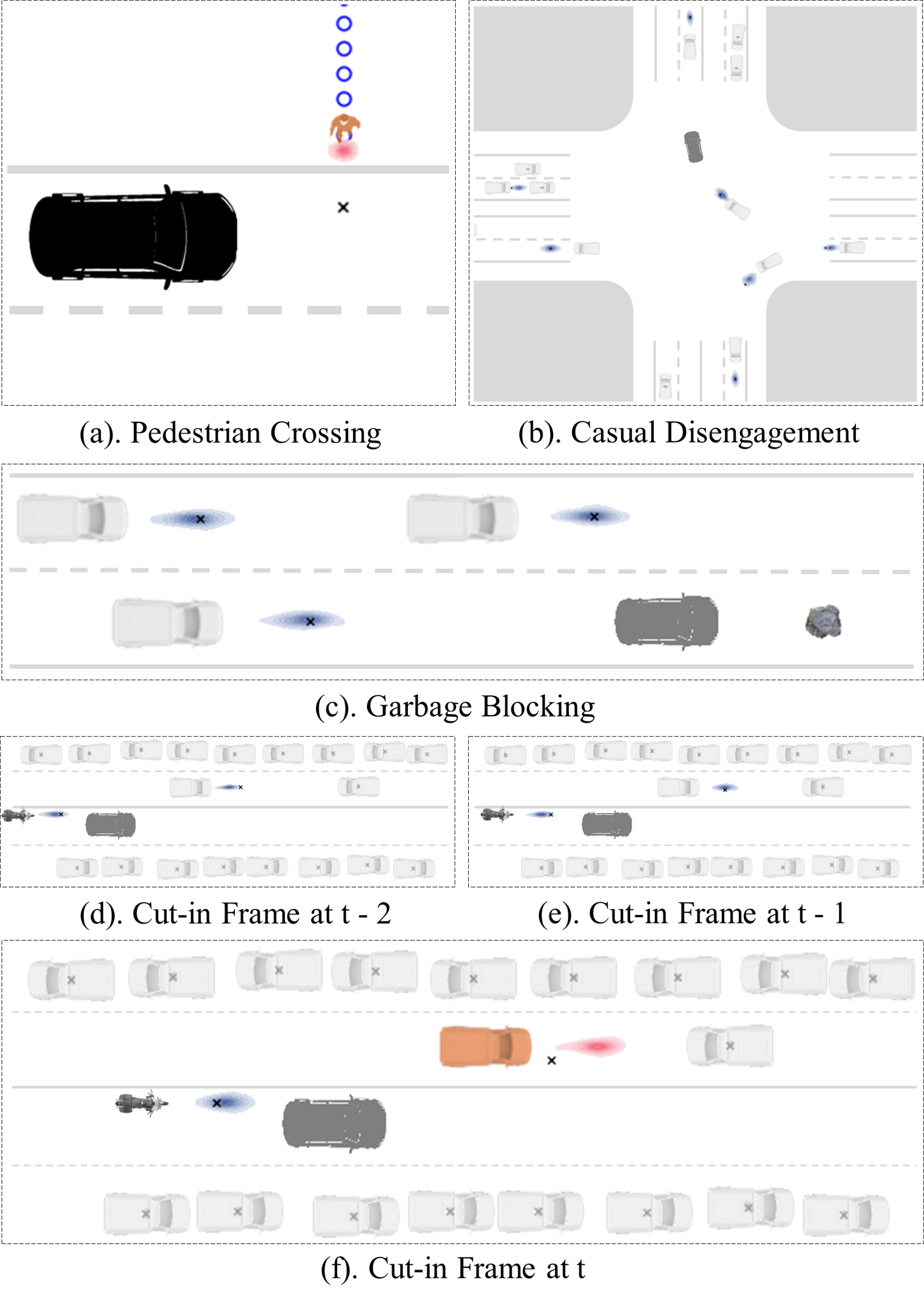}
\caption{Examples of Disengagement Reason Detection. The colored region in front of each object represents the future trajectory distribution predicted by the CVAE model, while the x-dots indicate the actual arrival positions of the objects. (a) A pedestrian unexpectedly accelerates, (b) Objects behave as expected, indicating a casual disengagement, (c) Non-policy-failure cases where objects act as anticipated, and (d)(e)(f) A vehicle unexpectedly performs a cut-in maneuver at frame \textit{t}. }
\label{fig:reasoning}
% \vspace{-5mm}
\end{figure}

% This part changes to Didi's scenario
For the first pedestrian crossing scenario, the result is shown in Fig. \ref{fig:reasoning}(a).
Historical state data indicate that the pedestrian was expected to yield; however, the pedestrian unexpectedly accelerated instead. Consequently, the start of the disengagement reason is identified as the moment when the pedestrian begins to accelerate. 
In the cut-in scenario, which involves multiple parked and moving objects, disengagement reasoning identifies most objects as non-contributory, except for the cut-in vehicle. Based on the historical state distribution, the cut-in vehicle was expected to remain in its lane, but it unexpectedly initiated an overtaking maneuver, as shown in Fig. \ref{fig:reasoning}(d) to \ref{fig:reasoning}(f). Thus, the start of the disengagement reason is pinpointed to the moment when the vehicle begins to overtake. These cases demonstrate that the proposed method effectively locates the object responsible for disengagement in complex scenarios and identifies the specific behavior causing it.

For the casual disengagement identification problem, the proposed method identifies 2221 frames as casual disengagements out of 2400, resulting in a successful identification rate ($\epsilon$) of 92.5\%. By introducing a reasonable assumption—that only nearby objects within 20 meters can cause disengagement—the number of mis-recognized disengagement frames reduces to 7 increasing $\epsilon$  to 99.7\%. An example of a frame identified as a casual disengagement is shown in Fig. \ref{fig:reasoning}(b), where the autonomous vehicle (AV) is navigating a complex junction.

For non-policy-failure disengagements, the method successfully identifies 568 frames out of 600, achieving a success rate of 94.7\%. A representative example is illustrated in Fig. \ref{fig:reasoning} (c).

\begin{table*}[!t]
\caption{Comparison of DRARL Performance with Baselines}
\centering

\label{tab:my-table}
\small % 缩小字体以适应宽度
\begin{tabularx}{0.85\textwidth}{@{}l *{2}{ccc}@{}} % 使用 tabularx 和 siunitx 对齐数字
\toprule
\multirow{2}{*}{Method} & 
\multicolumn{3}{c}{Scenario 1} & 
\multicolumn{3}{c}{Scenario 2} \\ 
\cmidrule(lr){2-4} \cmidrule(lr){5-7} % 用 cmidrule 分隔场景
& {Avg. Reward} & {Collision Rate} & {Pass Rate} & {Avg. Reward} & {Collision Rate} & {Pass Rate} \\ 
\midrule
SAC     & -0.1839 & 45\% & 55\% & -0.2137  & 42\%  & 58\%  \\
Randomly Reason RL & 0.0532  & 12\% & 49\% & -0.3235  & 35\%  & 50\%  \\
\textbf{DRARL (Proposed)} & \textbf{0.6661}  & \textbf{0\%} & \textbf{82\%}  & \textbf{0.6088} & \textbf{6\%} & \textbf{81\%}  \\
\bottomrule
\end{tabularx}
\label{table:overall_performance}
\end{table*}

\subsection{Policy Improvement Result in Disengagement Cases}

The policy improvement performance in disengagement cases is shown in Table.\ref{table:overall_performance}. As mention above, we randomly select 5\% of disengagement cases in each type of scenario for training. And the trained policy will be tested in all cases. 

The proposed DRARL outperforms baselines in both scenarios, achieving the highest average reward and scenario pass rate. The SAC policy, trained directly on disengagement logs, can only pass slightly more than half of the test cases that resemble the original training scenarios. This underscores the limitations of training with a single data point, as the policy may overfit to specific patterns and fail to generalize effectively. Another baseline, Randomly Reason RL,  results in an even lower scenario pass rate than SAC, highlighting the critical importance of accurately identifying the disengagement reason. Furthermore, the Randomly Reason RL policy tends to be overly conservative, as evidenced by its low collision rate but frequent instances of the ego vehicle becoming stuck in 39\% of test cases in Scenario 1 and 15\% in Scenario 2. 

\begin{figure}[h]
\centering
\includegraphics[width=0.9\linewidth]{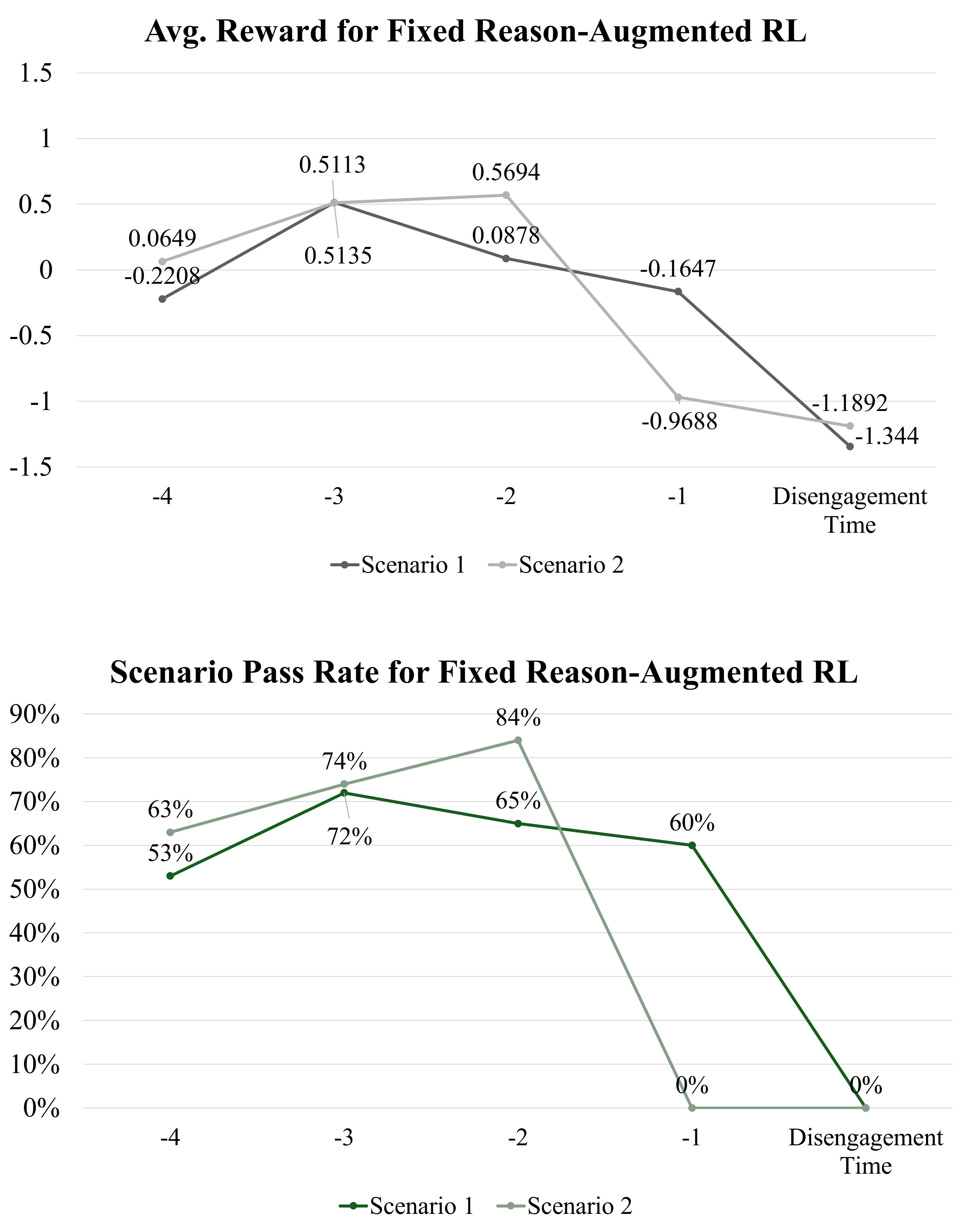}
\caption{The average reward and pass rate of the fixed reason-augmented reinforcement learning baseline. }
\label{fig:frarl_result}
% \vspace{-5mm}
\end{figure}

To further demonstrate the impact of the reason of disengagement's impart on policy training improvement, we compared the performance of different fixed-reason augmented RL methods, as illustrated in Figure \ref{fig:frarl_result}. As previously mentioned, these methods are similar to DRARL but utilize manually fixed disengagement reason moments during training instead of inferred by the model based on out-of-distribution (OOD) states. We tested four different fixed reason moments. The results reveal that the choice of the disengagement reason moment significantly influences the policy training effectiveness. If we choose the disengagement moment itself  as the reason (i.e., the Imagination Environment is constructed from this moment), the trained policy exhibits almost no performance improvement. This is because the root cause of the disengagement case (i.e., the potential policy failure) typically occurs before the moment, and thus, policy improvement must begin at an earlier stage. Moreover, the earlier the chosen moment doesn't means the greater the performance enhancement. In our experiments, DRARL outperforms the baseline methods using fixed moments, achieving higher average rewards and scenario pass rates, which clearly validates the effectiveness of our approach.

\section{Conclusion}

This work proposes the  disengagement-reason-augmented reinforcement learning (DRARL) method. DRARL efficiently enhances the policy while avoiding unnecessary updates triggered by incidental disengagement cases or overly conservative policy adjustments.
We reconstruct four types of disengagement cases collected from real-world autonomous driving road test with different reasons to evaluate our work.
The results show that the proposed DRARL method can find the reasons causing disengagement, including the specific agent and its out-of-distribution state.
Furthermore, by leveraging the reason-augmented imagination environment, DRARL improves the policy, enabling it to not only handle the original disengagement case but also generalize to similar cases. 

In general, the goal of this research is to achieve automatic and continual improvement of the autonomous driving policy during on-road test, paving the way for the development of fully autonomous vehicles. .

% \addtolength{\textheight}{-12cm}   % This command serves to balance the column lengths
                                  % on the last page of the document manually. It shortens
                                  % the textheight of the last page by a suitable amount.
                                  % This command does not take effect until the next page
                                  % so it should come on the page before the last. Make
                                  % sure that you do not shorten the textheight too much.

%%%%%%%%%%%%%%%%%%%%%%%%%%%%%%%%%%%%%%%%%%%%%%%%%%%%%%%%%%%%%%%%%%%%%%%%%%%%%%%%

%%%%%%%%%%%%%%%%%%%%%%%%%%%%%%%%%%%%%%%%%%%%%%%%%%%%%%%%%%%%%%%%%%%%%%%%%%%%%%%%

%%%%%%%%%%%%%%%%%%%%%%%%%%%%%%%%%%%%%%%%%%%%%%%%%%%%%%%%%%%%%%%%%%%%%%%%%%%%%%%%

\bibliographystyle{IEEEtran}
\bibliography{reference}

% \section*{APPENDIX}

% \subsection{Detailed Testing Result}

% \begin{figure}[!t]
% \centering
% \includegraphics[width=\linewidth]{fig/test_result.png}
% \caption{Detailed Testing Result.}
% \label{fig:test_result}
% \end{figure}

% The detailed testing result is shown in Fig. \ref{fig:test_result}.
% For Group 1, the parameter means (longitudinal relative distance in meter, the lateral displacement in meter).
% For Group 2, the parameter means (longitudinal relative cut-in distances in meters, the absolute cut-in speed in kilometers/second).
% The case with $*$ is the original case.
% $Y$ denotes the case is passed and $N$ denotes the case is failed.

% \section*{ACKNOWLEDGMENT}

% The preferred spelling

%%%%%%%%%%%%%%%%%%%%%%%%%%%%%%%%%%%%%%%%%%%%%%%%%%%%%%%%%%%%%%%%%%%%%%%%%%%%%%%%

\end{document}